\documentclass[11pt]{article}

\usepackage[margin=1in]{geometry}
\usepackage{mathptmx}
\usepackage[scaled=0.92]{helvet}
\usepackage{amsmath,amssymb}
\usepackage{booktabs}
\usepackage{makecell}
\usepackage{graphicx}
\usepackage{xcolor}
\usepackage{caption}
\usepackage{microtype}
\usepackage{enumitem}
\usepackage[colorlinks=true,linkcolor=blue!55!black,citecolor=blue!55!black,urlcolor=blue!55!black]{hyperref}

\hypersetup{pdftitle={The Harness Effect: How Orchestration Design Sets the Token Economics of Enterprise Agentic AI}, pdfauthor={Writer, Inc.}}

\newcommand{\NP}{22}                    
\newcommand{\NM}{six}                   
\newcommand{\CostA}{\$0.21}\newcommand{\CostB}{\$0.12}\newcommand{\CostD}{41\%}
\newcommand{\LatA}{48\,s}\newcommand{\LatB}{27\,s}\newcommand{\LatD}{44\%}
\newcommand{\SpeedX}{1.8$\times$}
\newcommand{\TokA}{14.2k}\newcommand{\TokB}{8.8k}\newcommand{\TokD}{38\%}
\newcommand{\QA}{0.78}\newcommand{\QB}{0.81}
\newcommand{\QPDgain}{82\%}
\newcommand{\CPMA}{54.9}\newcommand{\CPMB}{92.0}\newcommand{\CPMgain}{68\%}
\newcommand{\CostRange}{33\% to 61\%}
\newcommand{\FreezeDate}{2026-06-07}

\title{\LARGE\bf The Harness Effect: How Orchestration Design Sets\\[2pt] the Token Economics of Enterprise Agentic AI}

\author{\parbox{0.94\textwidth}{\centering
Muayad~Sayed~Ali, Aliaksandra~Novik, Anji~Boddupally, Artem~Yavorskyi, Chris~Nickerson, Daniel~Rica, Emily~DuGranrut, Felix~Leung, Garrett~Prince, Grace~Barnett, Heath~Robinson, Hosain~Al~Ahmad, Jesse~Resnick, Jyothi~Swaroop~Meruga, Leonid~Kuznetsov, Brock~Perry, Luke~Gorham, Marie~Schmoll, Michael~Paciullo, Saumya~Das, Sharath~Sheripally, Tommy~Griscom, Mykyta~Osadchyi, Neha~Mantri, Nick~Westrum, Olivia~Benowitz, Parikshith~Kulkarni, Radik~Chernyshov, Rakshith~Vasudev, Rohith~Nadimpally, Vikas~Gangadevi, and Waseem~AlShikh\\[9pt]
Writer, Inc.\\[3pt]
{\small\texttt{\{muayad, aliaksandra, anji, artem, chris, daniel, emily, felix, garrett, grace, heath, hosain, jesse, jyothi, leonid, brock, luke, marie, michael, saumya, sharath, tommy, mykyta, neha, nick, olivia, parikshith, radik, rakshith, rohith, vikas, waseem\}@writer.com}}
}}

\date{October 2026}

\begin{document}
\maketitle

\begin{abstract}
Agentic systems in enterprise deployments rarely answer a request with a single model call. They execute a loop of calls whose token volume is determined largely by the orchestration software that assembles context, exposes tools, sequences turns, delegates work, and records what was spent. We refer to this layer as the harness. This paper examines the token economics of the harness and introduces the notion of \emph{token maxing}, a development trajectory in which quality is purchased with growing token intensity at a declining marginal return. We argue that such a trajectory originates at the orchestration layer rather than in the model, and that it is also most effectively corrected there. To isolate the effect of the orchestration layer, we conduct a controlled swap: \NP{} locked evaluation tasks are executed on \NM{} foundation models (Claude Sonnet~4.6, Gemini~3.1, Gemini Flash~3.5, Qwen~3.6, GLM~5.1, and Palmyra~X6) under two orchestration layers, a conventional production agent loop that serves as a frozen baseline and the Writer Agent Harness, with tasks, models, judges, and price tables held fixed. Under the harness, blended cost per task falls by \CostD{} (\CostA{} to \CostB{}), median wall-clock time by \LatD{} (\LatA{} to \LatB{}), and tokens per task by \TokD{} (\TokA{} to \TokB{}), while task-completion quality is unchanged within the resolution of the sample (\QA{} to \QB{}). The efficiency gains are model-invariant, with every model's cost falling by \CostRange{}, whereas the quality gains depend on model capability: the mean improvement a model obtains from the harness is strongly correlated with its baseline score ($r=0.99$, $n=6$), a relationship we term \emph{harness leverage}. Quality per dollar rises by \QPDgain{}, and task completions per million tokens rise from \CPMA{} to \CPMB{}. On this workload, the choice of orchestration layer altered cost per task by more than the choice between the least and the most expensive model. We formalize the cost of an agentic task at the orchestration layer, including an effective input price under prompt caching; describe six families of mechanisms by which the harness reduces that cost; and compare six widely used agent systems against the same criteria.
\end{abstract}

\section{Introduction}
\label{sec:intro}

An agentic task is rarely a single model call. A request such as ``reconcile these two contracts and draft the redline memo'' typically unfolds into a dozen or more turns, each of which submits a system prompt, tool schemas, retrieval payloads, intermediate reasoning, and tool outputs, and each of which, in a naive implementation, resubmits everything that preceded it. The token cost of the task is the sum over this loop, and the loop is controlled not by the model but by the software around it. We use the term \emph{harness} for that software: the orchestration layer that determines what enters the context window, which tools are visible, when retrieval occurs, when a call is retried, when work is delegated, and when the loop terminates.

The prevailing response to rising capability requirements has been to spend more tokens. Reasoning models emit thousands of deliberation tokens per answer; agent frameworks replay conversation history, so that cumulative input grows quadratically in the number of turns; tool ecosystems inject every available schema into every call. Per-token prices have fallen steadily \cite{epoch2025}, and the falling unit price has financed this pattern: tokens are treated as nearly free at the margin, and consumption scales accordingly. The dynamic is a familiar one. Jevons observed that improvements in the efficiency of coal use raised total coal consumption rather than lowering it \cite{jevons1865}, and the same logic applies to inference. The resulting development trajectory, in which quality is purchased with monotonically growing token intensity at a declining marginal return, is what we call \emph{token maxing} (Definition~1, Section~\ref{sec:framing}). It does not appear in benchmark tables, which report quality; it appears in invoices, which report tokens.

Most published work on inference efficiency operates within a single model call, through prompt compression \cite{llmlingua}, budget-constrained reasoning \cite{tale}, terse decoding \cite{cod}, speculative decoding \cite{spec}, or serving-side memory management \cite{vllm}, or between models, through routing and cascading \cite{frugalgpt,routellm}. Both lines of work take the orchestration layer as given. Yet the harness governs every term of the token bill except the model's own verbosity. Whether the system prompt is replayed or cached, whether history is replayed or compacted, whether tool schemas are broadcast or disclosed on demand, how large retrieval payloads are, and when retries occur are all decisions made in orchestration code. If the harness is the layer that composes model calls into completed work, it is also the layer that sets the price of that work.

This paper tests that claim directly, using an experiment that the layered architecture makes possible: the tasks and the models are held constant and only the orchestration layer is changed. We evaluate \NP{} locked, capability-audited enterprise tasks on \NM{} foundation models spanning five vendors and three weight classes, under two orchestration layers: a conventional production agent loop, frozen on \FreezeDate{} and used as the baseline, and the Writer Agent Harness, a model-agnostic orchestration layer with sub-agent delegation. Tasks, prompts, model identifiers, judges, and price tables are identical across the two arms; the only variable is the orchestration code.

\medskip
\noindent The principal findings (Section~\ref{sec:results}) are the following.

\begin{enumerate}[leftmargin=1.6em,itemsep=2pt]
\item \emph{Efficiency improves uniformly.} Blended across models, cost per task falls by \CostD{} (\CostA{} to \CostB{}), median wall-clock time by \LatD{} (\LatA{} to \LatB{}, a \SpeedX{} speed-up), and tokens per task by \TokD{} (\TokA{} to \TokB{}). The per-model cost reduction ranges from $-$33\% (Gemini~3.1) to $-$61\% (Flash~3.5); no model is exempt. On this workload the orchestration layer is a larger cost lever than model selection: moving from the most to the least expensive model under the baseline saves 36\%, whereas retaining any model and adopting the harness saves \CostRange{}.
\item \emph{Quality is unchanged in aggregate and varies with the model at the margin.} Headline task completion moves from \QA{} to \QB{}, a difference we treat as parity at $n=\NP$. Of the 48 capability$\times$model cells, 30 improve, 11 are unchanged, and 7 regress; all 7 regressions occur on the three smaller models and are concentrated in orchestration-intensive capabilities (tool use through MCP, multi-step Playbooks). The mean quality gain a model obtains from the harness is strongly correlated with its baseline score ($r=0.99$): stronger models convert orchestration structure into quality, while weaker models can be burdened by it. We refer to this relationship as \emph{harness leverage} (Section~\ref{sec:leverage}).
\item \emph{Efficiency and capability are not traded against each other.} Quality per dollar rises by \QPDgain{}, and task completions per million tokens rise from \CPMA{} to \CPMB{} ($+$\CPMgain{}). The harness also introduces one capability absent from the baseline, delegation to sub-agents, which reaches a usable level of reliability only on the two strongest models (0.85--0.86); this is a concrete instance of the capability floor that orchestration features carry.
\end{enumerate}

\medskip
\noindent\textbf{Contributions.}
(1)~A formal account of token economics at the orchestration layer, comprising a decomposition of per-task token spend into harness-controlled terms, an effective input price under prompt caching, and a definition of token maxing as a measurable development trajectory (Section~\ref{sec:framing}).
(2)~A controlled methodology that isolates orchestration effects from model effects by means of a locked task set, a frozen baseline, and trace-level accounting (Section~\ref{sec:setup}).
(3)~Empirical results across \NM{} models: model-invariant efficiency gains, capability-dependent quality gains (harness leverage), and a capability floor for advanced orchestration features (Section~\ref{sec:results}).
(4)~An inventory of the harness mechanisms (prefix stability and cache placement, structured compaction, context offload, durable suspension, failure-spend governance, and model-agnostic invariants), mapped onto the cost decomposition, together with an architectural comparison of six widely used agent systems against the same criteria (Sections~\ref{sec:mechanisms}--\ref{sec:landscape}).
(5)~An economic analysis of why savings at the orchestration layer compound, applying to every model, every vendor migration, and every unit of volume, and of the implications for the decision to build or to rent orchestration infrastructure (Section~\ref{sec:discussion}).

\section{Related Work}
\label{sec:related}

\paragraph{Model- and serving-side efficiency.} A substantial literature reduces the cost of processing a given sequence of tokens. Speculative decoding accelerates autoregressive generation without changing the output distribution \cite{spec}, and PagedAttention and related memory managers raise serving throughput \cite{vllm}. These techniques lower the price of a token; they do not change how many tokens an agent consumes, which is the quantity of interest here.

\paragraph{Prompt-side token reduction.} LLMLingua and its successors compress prompts while preserving task performance \cite{llmlingua}; token-budget-aware reasoning constrains the length of deliberation \cite{tale}; Chain-of-Draft shows that terse intermediate reasoning can match verbose chain-of-thought \cite{cod,cot}. Each of these operates within a single call. An agent harness issues such calls many times per task and introduces cross-call structure (history replay, tool schemas, retrieval) that no single-call method observes. Our results suggest that the larger savings lie in that cross-call structure: tokens per task fall by \TokD{} with no change to models or prompts.

\paragraph{Routing and cascades.} FrugalGPT and RouteLLM reduce cost by directing easier queries to cheaper models \cite{frugalgpt,routellm}. Routing determines which model is billed; the harness determines the size of the bill for whichever model is selected. The two are complementary, and Section~\ref{sec:codesign} argues that the capability-floor result reported here bears on routing: requests should be assigned not only by difficulty but by the orchestration features they will exercise.

\paragraph{Agent architectures.} ReAct interleaves reasoning with tool calls \cite{react}; Reflexion adds verbal self-correction \cite{reflexion}; Toolformer trains tool invocation into the model \cite{toolformer}; SWE-agent demonstrates that the design of the agent--computer interface materially affects task success \cite{sweagent}; MemGPT introduces operating-system-style context paging \cite{memgpt}; and the Model Context Protocol standardizes tool connectivity \cite{mcp}. This line of work established that scaffolding shapes capability. Studies of long-context behavior supply one mechanism: models attend unevenly across long inputs, so an inflated context can reduce accuracy while raising cost \cite{lost}. We extend the observation that scaffolding shapes capability with its economic counterpart, namely that scaffolding sets cost, and quantify both under a controlled swap.

\paragraph{Evaluation and test-time compute.} LLM-as-judge protocols \cite{judge} and locked benchmark suites \cite{swebench} inform our measurement design. Work on the optimal allocation of test-time compute \cite{snell} shows that quality can be purchased with inference tokens; we treat that purchase as an economic decision with a declining marginal rate, and the harness as the agent making it. Reflective prompt-evolution systems such as GEPA \cite{gepa} optimize the instructions a harness issues and are natural complements to the layer studied here.

\paragraph{Economics of inference.} Scaling laws priced training compute \cite{kaplan,chinchilla}, and Epoch~AI documents declines of several orders of magnitude in inference prices \cite{epoch2025}. Jevons's observation in 1865 that more efficient use of coal raised its total consumption \cite{jevons1865} is the relevant precedent; Section~\ref{sec:framing} argues that the orchestration layer is where consumption discipline can in practice be implemented.

\paragraph{The harness as an object of study.} Closest to the present work, a recent line of research treats the harness as a first-class object. Gu argues that progress in agentic systems is now a matter of system scaling (context governance, memory, skill routing) as much as of model scaling \cite{harnessscaling}, and Harness-Bench reports, over 5{,}194 trajectories, that completion rate, efficiency, and failure behavior vary substantially across model--harness configurations \cite{harnessbench}. Practitioner reports point in the same direction: guidance on context engineering for production agents treats the KV-cache hit rate as the primary metric \cite{manus}; provider documentation prices cached input at roughly one tenth of the list rate \cite{promptcache}; controlled measurements show quality degrading as input length grows \cite{contextrot}; and Anthropic reports token consumption of roughly $4\times$ chat levels for single agents and $15\times$ for multi-agent systems \cite{anthmulti}. The present paper adds the economic layer that this literature lacks: a formal account of token cost at the harness (Section~\ref{sec:framing}), the mechanisms that implement it (Section~\ref{sec:mechanisms}), and a controlled experiment that prices it (Section~\ref{sec:results}).

\section{Token Economics at the Orchestration Layer}
\label{sec:framing}

\subsection{Cost of an agentic task}

Consider a task executed as a $k$-turn agent loop. Turn $i$ submits input tokens $T^{\mathrm{in}}_i$ and receives output tokens $T^{\mathrm{out}}_i$. With input price $p_{\mathrm{in}}$ and output price $p_{\mathrm{out}}$ per token, the task costs
\begin{equation}
C \;=\; \sum_{i=1}^{k}\Big( p_{\mathrm{in}}\,T^{\mathrm{in}}_i + p_{\mathrm{out}}\,T^{\mathrm{out}}_i \Big).
\label{eq:cost}
\end{equation}

\begin{figure}[t]
\centering
\includegraphics[width=0.62\linewidth]{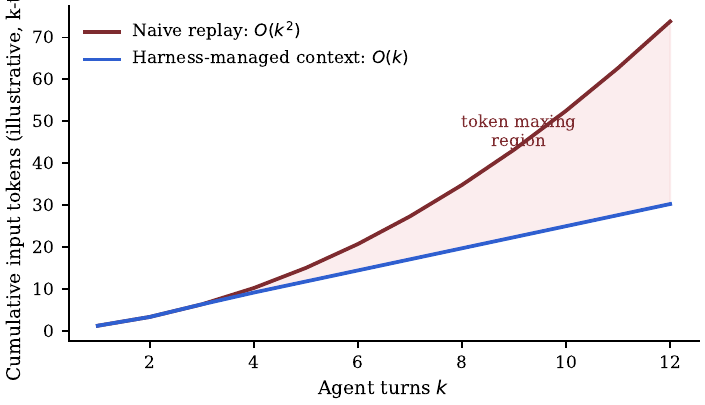}
\caption{Cumulative input tokens for a $k$-turn agent loop under full-history replay, which grows as $O(k^2)$ (Eq.~\ref{eq:quadratic}), and under harness-managed context (prefix caching, history compaction, offloaded tool outputs), which grows as $O(k)$. The shaded region represents spend that purchases no additional quality. The figure is schematic; measured aggregates are reported in Section~\ref{sec:results}.}
\label{fig:replay}
\end{figure}

The input side decomposes into terms that the harness constructs:
\begin{equation}
T^{\mathrm{in}}_i \;=\; \underbrace{S_i}_{\text{system}} + \underbrace{H_i}_{\text{history}} + \underbrace{G_i}_{\text{tool schemas}} + \underbrace{R_i}_{\text{retrieval}} + \underbrace{U_i}_{\text{user turn}},
\label{eq:decomp}
\end{equation}
and the loop multiplies every term by retries and dead-end branches, which the harness also governs. A naive harness replays the full transcript, so that $H_i = \sum_{j<i}(U_j + T^{\mathrm{out}}_j + O_j)$, where $O_j$ is the tool output at turn $j$. Total input tokens then grow quadratically in the number of turns,
\begin{equation}
\sum_{i=1}^{k} T^{\mathrm{in}}_i \;\approx\; kS \;+\; \frac{k(k-1)}{2}\,\bar m \;+\; k\bar G \;+\; \textstyle\sum_i R_i,
\label{eq:quadratic}
\end{equation}
with $\bar m$ the mean per-turn payload. A harness that compacts history, caches the invariant prefix (tool schemas included), offloads bulky tool outputs, and limits retrieval to the evidential minimum reduces the quadratic term to approximately linear growth (Figure~\ref{fig:replay}). The model is unchanged; only the composition of its input differs.

Two further considerations refine Eq.~\eqref{eq:cost}. First, agent workloads are input-dominated. Because the transcript is resubmitted on every turn, production agents report input-to-output token ratios near 100:1 \cite{manus}, so the $p_{\mathrm{in}}$ term accounts for nearly the whole bill. Second, the price of an input token is not a single number. Providers serve tokens that repeat a previously seen prompt prefix from cache at a substantial discount, roughly $0.1\times$ the base input rate \cite{promptcache,manus}. If a fraction $h$ of input tokens are cache reads billed at multiplier $\kappa$, the effective input price is
\begin{equation}
p^{\mathrm{eff}}_{\mathrm{in}} \;=\; p_{\mathrm{in}}\big(1 - h\,(1-\kappa)\big), \qquad \kappa \approx 0.1,
\label{eq:cache}
\end{equation}
so a harness that keeps $h$ near $1$ pays roughly one tenth of the list price for the dominant term of Eq.~\eqref{eq:cost}. The quantity $h$ is a property neither of the model nor of the provider; it is determined by the byte-level stability of the prompt across turns, which is set by how the orchestration layer assembles context. The harness therefore controls both factors of the bill: the number of tokens submitted (Eq.~\eqref{eq:decomp}) and the price at which the dominant ones are billed (Eq.~\eqref{eq:cache}).

\subsection{Efficiency metrics and a definition of token maxing}

Define \emph{token intensity} $\tau$ as the total number of tokens per completed task, and let $Q\in[0,1]$ denote task quality under a fixed judge protocol. We use two efficiency metrics:
\begin{equation}
\eta_{\$} \;=\; \frac{Q}{C} \quad \text{(quality per dollar)}, \qquad
\mathrm{CPM} \;=\; \frac{Q \cdot 10^{6}}{\tau} \quad \text{(task-completions per million tokens)}.
\label{eq:eff}
\end{equation}

\begin{quote}
\textbf{Definition 1 (Token maxing).}\label{def:maxing}
A development trajectory $\{(Q_t,\tau_t)\}_t$ exhibits \emph{token maxing} if token intensity grows, $\tau_{t+1} > \tau_t$, while marginal quality per token declines, $\frac{Q_{t+1}-Q_t}{\tau_{t+1}-\tau_t} < \frac{Q_t}{\tau_t}$ --- i.e., each release buys quality at a worse token exchange rate than the system's running average.
\end{quote}

Token maxing is individually rational for a team evaluated on benchmark quality and collectively costly for the organization that pays per token. It is also self-reinforcing under falling prices: when $p$ falls, the budget-constrained optimum shifts toward higher $\tau$, and total spend $N\cdot C$ can rise even as $p\cdot\tau$ per task falls, which is Jevons's argument restated for tokens \cite{jevons1865,epoch2025}. The remedy is not a cheaper token but a higher $\mathrm{CPM}$, that is, the same work completed with fewer tokens. Since four of the five input terms in Eq.~\eqref{eq:decomp}, together with the retry multiplier, are determined by code rather than by the model, that remedy is available at the orchestration layer.

\subsection{Compounding at fleet scale}

Consider a fleet that runs $N$ tasks per month across models $m\in\mathcal{M}$ with mix weights $w_m$, so that monthly spend is $\sum_m w_m N\, C_m$. A model-side optimization improves a single $C_m$, and a routing policy improves the mix $w$. An improvement to the harness multiplies every $C_m$ by a factor $(1-s_m)$ simultaneously, and continues to do so when the model set changes, because it is implemented above the model API. Empirically (Section~\ref{sec:invariance}), $s_m \in [0.33, 0.61]$ for all \NM{} models in this study. This property distinguishes the harness from other components of the stack: its returns do not depend on which vendor's model is in use.

\section{The Writer Agent Harness}
\label{sec:harness}

\subsection{Architecture}

In Writer's architecture the harness is the runtime situated between the application and the foundation model. It owns (i)~\emph{context assembly}: the system prompt, conversation state, and retrieval payloads from document and knowledge-graph grounding; (ii)~the \emph{tool layer}: native tools and external connectors via the Model Context Protocol \cite{mcp}, including schema exposure and the mediation of calls; (iii)~\emph{workflow execution}: multi-step Playbooks executed end to end; (iv)~\emph{delegation}: spawning scoped sub-agents and merging their results, a capability absent from the baseline loop; and (v)~\emph{observability}: a trace shim that records prompt tokens, completion tokens, tool events, and wall-clock time (\texttt{duration\_seconds}) for every turn of every run. The trace shim is what makes the accounting in this paper possible. Cost is computed when the report is built, by applying a pinned public price table to the recorded token counts, so that both arms are priced identically. The same component serves as the enterprise control plane: the trace shim that meters tokens is also the audit trail, the progressive disclosure of tools that reduces tokens is also the mechanism of tool governance, and deterministic workflow execution is what makes agent behavior reviewable. Efficiency and control are therefore properties of a single component, which is why the harness is central to the rest of the platform.

The harness is model-agnostic by construction. The executing model is a configuration value (\texttt{model\_name}), which is what permits the controlled swap of Section~\ref{sec:setup}, in which six models run under two orchestration layers against one task set and one judge panel. The harness is the runtime behind Writer's Action Agent, whose published results (61\% on the hardest GAIA level and the highest overall score on the CUB computer-use benchmark) characterize the production system that this evaluation instruments \cite{actionagent}.

\subsection{Comparison with a conventional agent loop}

The baseline is the agent loop previously deployed in production, frozen on \FreezeDate{}. Its design is representative of common practice, and we state it concretely because each element constitutes a token-economics decision made by default: a monolithic system prompt of roughly 49\,KB replayed on every turn; tool invocation parsed by regular expressions from XML emitted in the text stream; destructive middle-truncation when the context overflows; per-model prompt tuning; and waiting implemented by polling. The Writer Agent Harness, evaluated at its intended general-availability configuration, replaces each of these with an explicit contract: a single execution path for all six models; native tool calling as the only invocation path (the XML path was removed rather than ported); non-destructive structured compaction in place of truncation; early streaming with live tool-call status; cancellation and retry as first-class states; and sub-agents, which delegate a sub-task to one or more spawned agents with scoped context and merge their results, a capability the baseline does not possess.

\subsection{Mechanisms}
\label{sec:mechanisms}

Mapping the observed differences onto Eqs.~\eqref{eq:decomp} and \eqref{eq:cache}, the savings must arise from the terms that the harness rewrites. The design objective can be stated concisely: maximize the fraction of tokens that are cached, that are relevant to the current decision, and that are spent within committed and recoverable work, and enforce each property structurally rather than relying on model behavior. Six families of mechanisms implement this objective.

\begin{figure}[t]
\centering
\includegraphics[width=0.78\linewidth]{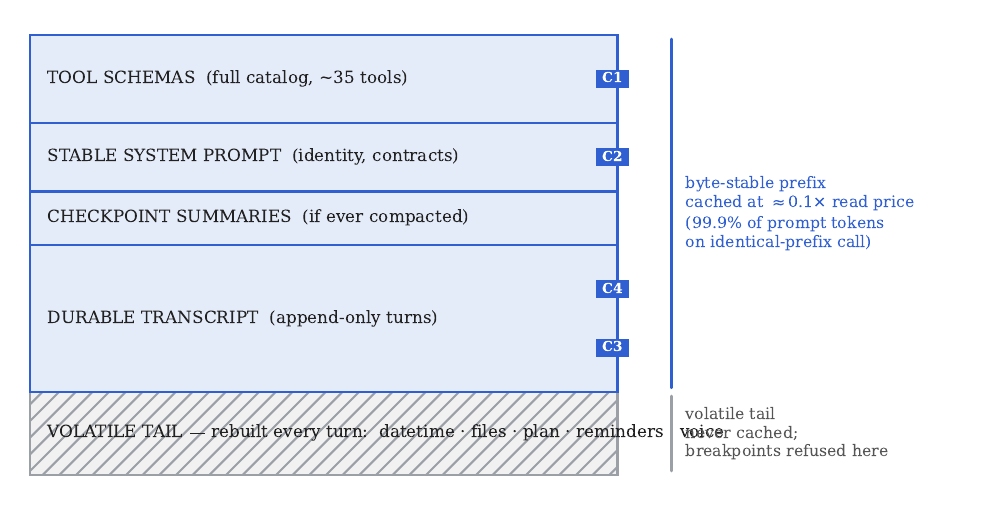}
\caption{Structure of the harness prompt. A byte-stable prefix (tool-schema catalog, stable system prompt, append-only transcript) carries up to four provider cache breakpoints (C1--C4); content that changes from turn to turn is confined to a volatile tail that is rebuilt on each turn and excluded from caching. On an identical-prefix call, 99.9\% of prompt tokens (7{,}876 of 7{,}886) were served as cache reads, which prices the dominant input term at approximately $0.1\times$ the list rate (Eq.~\ref{eq:cache}).}
\label{fig:prompt}
\end{figure}

\paragraph{(1) Prefix stability and cache placement.}
Every prompt the harness emits has a fixed physical structure (Figure~\ref{fig:prompt}). A byte-stable prefix contains the full tool-schema catalog, the stable system prompt, and the append-only durable transcript; a volatile tail, rebuilt on every turn, contains the clock, file listings, plan state, one-shot reminders, and voice and custom instructions. The separation is enforced as a correctness rule rather than treated as an optimization: content that changes between turns may not appear in the prefix, and the cache-marker logic refuses to place a breakpoint at or after the first volatile message. Up to four provider breakpoints are set, after the schema catalog, after the stable system prompt, and on the two most recent durable turns, with one-hour retention latched for the duration of a session so that the policy does not change mid-run. Measured on an identical-prefix call in the harness repository, 7{,}876 of 7{,}886 prompt tokens (99.9\%) were served as cache reads; by Eq.~\eqref{eq:cache} with $\kappa\approx0.1$, such a turn pays approximately one tenth of the list price for its dominant term. Because agent workloads are input-dominated, the cache hit rate is the single most consequential cost variable under the harness's control, and it is determined by the placement of $S_i$, $G_i$, and $H_i$ rather than by the model.

\paragraph{(2) Structured, incremental compaction.}
When history reaches 80\% of the model's input budget, older turns are folded into a typed checkpoint comprising four artifacts: durable memory (decisions, constraints, rejected approaches), an eight-section execution summary written for resumability (current state, files touched, errors, next steps), the user's requirements preserved verbatim, and references to active skills. A live tail of the 4--12 most recent messages, bounded at 30\% of the budget, is always retained verbatim. Each checkpoint folds the previous one forward, so the cost of compaction remains incremental; summarization is performed by a less expensive helper model outside the main loop; and an empty or degraded summary aborts the compaction rather than being persisted. If the prompt still exceeds the budget, a deterministic sequence (shrinking the tail, then middle-truncating with an explicit instruction to restate what matters) guarantees that a request can be sent. Compaction and caching are designed together. A summarizer that rewrote history on every turn would destroy the prefix stability on which Eq.~\eqref{eq:cache} depends; checkpoints are therefore stored as durable rows, and the rebuilt prompt becomes the new cacheable prefix. This is the mechanism that reduces Eq.~\eqref{eq:quadratic} from quadratic to approximately linear growth (Figure~\ref{fig:replay}) without the destructive truncation used by the baseline.

\paragraph{(3) Context offload.}
A third family keeps information available without keeping it in the context window, reducing $R_i$ and the tool-output share of $H_i$. Sub-agents act as context firewalls: a child agent performs extended reading or searching in its own context and returns a summary bounded at 8\,KB, with citations carried in a metadata sidecar that the parent model does not read; delegation is depth-limited and idempotent under retries, so delegated exploration cannot inflate the parent loop. Skills are disclosed progressively: the prompt carries only a table of names and descriptions, the full skill document is installed in the sandbox and read only on invocation (with the inline seed bounded at 20K characters), and compaction later removes skill bodies deterministically while retaining the reference. Large tool outputs are spilled to files: shell output beyond 20K characters is shown as a head-and-tail preview with the complete output written to a workspace file (accompanied by a notice that success must not be inferred from the preview), web fetches return 8K characters inline with the full page written to a file, and oversized file reads are rejected with guidance rather than silently truncated. The filesystem thus serves as unbounded memory while the context holds pointers. Plan and canvas state are event-sourced and projected once per turn as a compact rendering, with plan-tool results replaced by one-line acknowledgements so that state is never duplicated across the transcript; the per-turn rendering also restates the objective, which counteracts goal drift over long horizons \cite{manus}. Background-task results are delivered exactly once as an ephemeral reminder in the tail; loaded media are bounded (at most four images or 2\,MB in context, with older items evicted and replaced by a reload stub); and ancillary work such as telemetry, reasoning-trace titles, and the materialization of deliverables runs outside the main loop on helper models.

\paragraph{(4) Durable suspension.}
Waiting is implemented as a continuation rather than as a loop. When a run requires a human response, an approval, or the completion of a long background job, it is suspended durably at no token cost and resumed by an ingress event; no polling turns are issued. The same durability layer bounds catastrophic spend. Every event is written to a write-ahead log before it is streamed, crashed or preempted runs resume under generation fencing at the next sequence number, and tool results are persisted before they are displayed. A crash that loses a 40-turn run would otherwise require those 40 turns to be purchased again; under the harness, the run resumes from durable state.

\paragraph{(5) Failure-spend governance.}
Every failure is assigned to a typed class (rate limit, stall, timeout, malformed stream, provider outage, permanent) before any decision is taken, and only whitelisted classes fall through to the next provider in the route plan. A failure in mid-stream becomes a discarded attempt: the partial draft is cleared, and no side effects (tool executions or durable writes) can originate from it. This is the element that generic library fallbacks omit; streaming failover in orchestration libraries is documented to operate only before the first chunk has been emitted \cite{langgraph}. A circuit breaker halts a model that issues a byte-identical failing tool call three times, with cause-aware steering (changing the arguments versus backing off); truncated outputs and output-length limits terminate as typed states rather than silently; the loop is limited to 50 iterations, and tool parallelism to four. Retries, dead ends, and repetitive loops multiply the cost in Eq.~\eqref{eq:cost} in a way that no per-token discount offsets, and the harness bounds that multiplier.

\paragraph{(6) Model-agnostic invariants.}
The model to run, the providers over which to run it, and the fallback order are supplied as a typed route plan; the loop never branches on a model name, and every provider stream is normalized to a single chunk contract before the loop observes it. Native tool calling is the only invocation path, supported by schema hygiene for weaker models: \texttt{\$ref}s are inlined, double-encoded JSON arguments are recovered, framework internals are removed from validation errors, and overloaded tool schemas are split when weaker models misuse them. These invariants hold regardless of the executing model, which accounts structurally for the model-invariance reported in Section~\ref{sec:invariance} and for the regularity with which harness leverage (Section~\ref{sec:leverage}) tracks model capability: the harness fixes the floor, and the model determines the ceiling.

A common thread runs through the six families: token economy and output quality are not separate objectives. Long and distractor-dense contexts measurably degrade every frontier model tested \cite{contextrot,lost}, so a mechanism that removes stale or bulky tokens reduces the bill and improves the model's working set at the same time. The grounding results of Section~\ref{sec:cases} illustrate this in a single case.

Two implementation notes are recorded for completeness. Sub-agent delegation currently executes as blocking calls, and automatic whole-turn retry on output-length limits is designed but not yet deployed. The mechanisms described above are those active in the configuration under evaluation.

\subsection{Related systems}
\label{sec:landscape}

Recent work argues that the harness should be treated as a first-class object of design and evaluation \cite{harnessscaling}, and cross-harness measurement finds that agent capability is a property of the model--harness configuration rather than of the model alone \cite{harnessbench}. Our claim here is narrower and economic: among widely used agent systems, token economics is not a first-class, published contract. Table~\ref{tab:landscape} summarizes six systems against the mechanism families of Section~\ref{sec:mechanisms}, assessed from public documentation and from a source-level study of caching behavior conducted during the design of the harness.

\begin{table}[t]
\centering\footnotesize
\caption{Token-economic mechanisms in widely used agent systems, assessed from public documentation and a design-time source study rather than from head-to-head measurement (see Section~\ref{sec:threats}). ``app'' denotes a mechanism left to the application to build and budget.}
\label{tab:landscape}
\begin{tabular}{@{}llcccccc@{}}
\toprule
System & Deployment class & \makecell{Model-\\portable} & \makecell{Structural\\cache policy} & \makecell{Compaction\\contract} & \makecell{Firewalled\\delegation} & \makecell{Zero-token\\waits} & \makecell{Per-task\\accounting} \\
\midrule
Claude Code & single-user client & no & yes & yes & partial & no & no \\
Claude Cowork & single-user client & no & yes & yes & partial & no & no \\
LangGraph & library & yes & app & app & app & partial & no \\
CrewAI & framework & yes & app & partial & no & no & no \\
AutoGen / AG2 & framework & yes & app & app & no & partial & no \\
Hermes Agent & personal agent & yes & partial & yes & yes & partial & no \\
\midrule
\textbf{Writer Harness} & \textbf{multi-tenant runtime} & \textbf{yes} & \textbf{yes} & \textbf{yes} & \textbf{yes} & \textbf{yes} & \textbf{yes} \\
\bottomrule
\end{tabular}
\end{table}

\paragraph{Vendor-integrated clients.} Claude Code and Claude Cowork \cite{claudecode} are among the most sophisticated harnesses in wide deployment, and several of their patterns (cache-breakpoint latching, byte-stable prefixes, sub-agent task splitting) were adopted from or validated against them during the design of the harness studied here. The differences lie in deployment class and contract. Both are single-user, client-side tools bound to one model vendor; their internal token management is not exposed as a per-task accounting surface; and their economics do not transfer across models. They are designed to optimize a session for an individual user, whereas an enterprise runtime must meter a fleet.

\paragraph{Orchestration libraries.} LangGraph \cite{langgraph} provides graph orchestration, checkpointing, and interrupts as low-level primitives and leaves prompt-cache policy, compaction, context offload, and failure-spend governance to the application. Token economics therefore becomes a responsibility of the application team that the framework neither measures, enforces, nor reports, and the framework's own documentation acknowledges the streaming-failover limitation noted in Section~\ref{sec:mechanisms}.

\paragraph{Multi-agent conversation frameworks.} CrewAI \cite{crewai} and the AutoGen lineage \cite{autogen} (the original framework, now in maintenance mode, continued by the AG2 community fork and succeeded in Microsoft's stack by the Agent Framework) organize work as conversations among role-prompted agents that share a transcript. Shared-transcript multi-agency multiplies tokens by construction, since each participating agent re-reads the growing conversation and carries its own role preamble. The most informative public measurement is Anthropic's: single agents consume roughly $4\times$ the tokens of a chat interaction and multi-agent systems roughly $15\times$, with token volume accounting for about 80\% of the variance in performance on their research evaluation \cite{anthmulti}. This is Definition~1 realized as an architecture, with quality purchased through the multiplication of tokens. The purchase can be justified, and Anthropic notes that only high-value, parallelizable tasks warrant it; but none of these frameworks meters the multiplier per task, and none bounds delegated context in the manner of a capped-summary sub-agent contract.

\paragraph{Open personal harnesses.} Hermes Agent \cite{hermes} is a capable, model-agnostic open-source harness with isolated sub-agents. The design-time source study found, however, that it does not place its tool schemas within a cached prefix, forgoing the largest single discount available on an input-dominated workload (Eq.~\eqref{eq:cache}), and as a personal, single-user agent it publishes no per-task token or cost contract.

\paragraph{Summary.} The final column of Table~\ref{tab:landscape} states the central point. Without per-task token accounting at the orchestration layer, token maxing cannot be observed, and what cannot be observed cannot be managed. The trace shim of the Writer harness makes cost per task and CPM measurable for each release because the meter resides in the layer that spends the tokens, and the release-gate practice described in Section~\ref{sec:kpi} depends on it. We note the epistemic status of this comparison: Table~\ref{tab:landscape} compares published designs and documentation rather than measured runs, and configuration-level measurement of the token dimension across harnesses, in the manner of Harness-Bench \cite{harnessbench}, remains future work.

\section{Experimental Setup}
\label{sec:setup}

\subsection{Design}

The experiment is a paired comparison. Both arms execute the same \NP{} tasks on the same \NM{} models with the same judges and the same pinned price table; only the orchestration layer differs. The baseline arm is the conventional production loop, invoked through a \texttt{model\_name} override, run once and frozen on \FreezeDate{} as the reference. The harness arm is the Writer Agent Harness at its intended general-availability configuration. Because prompts, models, and judges are held fixed, any systematic difference in tokens, cost, latency, or quality between the arms is attributable to the orchestration layer.

\subsection{Task set}

The evaluation uses $n=\NP{}$ locked-baseline prompts, audited for capability coverage and for defensible pass/fail criteria before either arm was run. The set spans nine capability areas that reflect production use of an enterprise agent platform: Model \& System Awareness (MSA: identity, scope, safe refusal), Grounding \& Retrieval (GDR: document and knowledge-graph grounding, web search, citation), Content Generation (CNG), Playbooks (PLY: predefined multi-step workflows executed end to end), MCP tool use (MCP: discovering and calling external connectors), Presentations (PRN: slide generation, an early-stage capability), Voice (VOX: brand voice and custom instructions), Image Analysis \& Generation (IMG), and, under the harness only, Sub-agents (delegated work). Tasks include multi-turn cases (for example, a three-turn Medicare grounding dialogue) and long-horizon cases (multi-step research synthesis).

\subsection{Models}

\begin{table}[t]
\centering\small
\caption{Models under evaluation. Every model is run identically under both arms.}
\label{tab:models}
\begin{tabular}{@{}llll@{}}
\toprule
Model & Vendor & Class & Role in analysis \\
\midrule
Claude Sonnet 4.6 & Anthropic & frontier & strong generalist \\
Gemini 3.1 & Google DeepMind & frontier & strong generalist \\
Gemini Flash 3.5 & Google DeepMind & fast tier & low-cost generalist \\
Qwen 3.6 & Alibaba & open weight & candidate model \\
GLM 5.1 & Zhipu AI & open weight & candidate model \\
Palmyra X6 & Writer & enterprise & platform-native model \\
\bottomrule
\end{tabular}
\end{table}

Table~\ref{tab:models} lists the \NM{} models: two frontier generalists, one fast-tier model, two open-weight candidates, and Writer's platform-native Palmyra~X6. The spread is deliberate. Because the central claim concerns a layer above the model, it must be tested across weight classes and vendors rather than on a single flagship model.

\subsection{Metrics}

\emph{Quality.} Headline quality is task completion, scored by an LLM-judge panel against the audited pass/fail criteria \cite{judge}. Secondary judges score coherence, communication quality, and hallucination; these are reported per prompt and did not determine any decision. \emph{Tokens.} Prompt and completion tokens are recorded per turn by the trace shim; token intensity $\tau$ is the per-task total. \emph{Cost.} Computed when the report is built as $(\text{prompt tokens} \times p_{\mathrm{in}}) + (\text{completion tokens} \times p_{\mathrm{out}})$ using a pinned OpenRouter price table, identically for both arms. \emph{Latency.} Wall-clock time from first token to terminal event (\texttt{duration\_seconds}), median across runs. \emph{User experience.} A hand-scored qualitative rubric (streaming responsiveness, tool-call transparency, citation rendering, error and cancellation handling, multi-turn coherence), reported separately from the automated metrics.

\subsection{Statistical considerations}

At $n=\NP{}$, the quality differences are directional rather than statistically significant, and we treat them as such throughout: the headline difference of $+0.03$ is reported as parity rather than as improvement. The cost, token, and latency differences are large and uniform in sign across all \NP{} prompts and all \NM{} models, and are therefore decisive at this sample size in a way that the quality differences are not. Failures of candidate models on advanced features are expected and are scored rather than excluded; model-agnosticism is a claim about the execution path, not a claim of uniform capability.

\section{Results}
\label{sec:results}

\subsection{Aggregate results}

\begin{figure}[t]
\centering
\includegraphics[width=\linewidth]{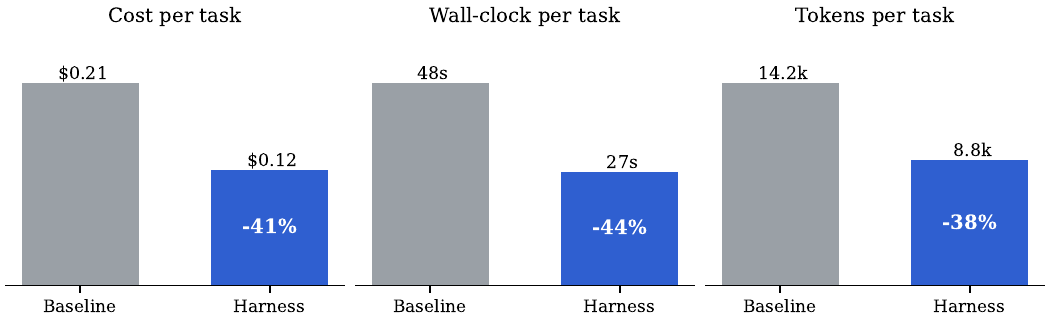}
\caption{Blended efficiency across \NM{} models and \NP{} tasks when the baseline loop is replaced by the harness with models held constant: cost per task $-$\CostD{}, median wall-clock $-$\LatD{}, tokens per task $-$\TokD{}.}
\label{fig:headline}
\end{figure}

\begin{table}[t]
\centering\small
\caption{Parity scorecard, blended across models. Quality is task-completion under the locked judge protocol; derived rows use Eq.~\eqref{eq:eff}.}
\label{tab:scorecard}
\begin{tabular}{@{}lccc l@{}}
\toprule
Dimension & Baseline & Harness & $\Delta$ & Reading \\
\midrule
Quality (task-completion) & \QA{} & \QB{} & $+0.03$ & parity at $n=\NP$ \\
Cost / task & \CostA{} & \CostB{} & $-\CostD$ & decisive \\
Wall-clock / task (median) & \LatA{} & \LatB{} & $-\LatD$ & decisive \\
Tokens / task & \TokA{} & \TokB{} & $-\TokD$ & decisive \\
\midrule
Quality per dollar ($\eta_{\$}$) & 3.71 & 6.75 & $+\QPDgain$ & derived \\
Completions per Mtok (CPM) & \CPMA{} & \CPMB{} & $+\CPMgain$ & derived \\
UI/UX (hand-scored) & baseline & improved & $\uparrow$ & qualitative \\
\bottomrule
\end{tabular}
\end{table}

Table~\ref{tab:scorecard} and Figure~\ref{fig:headline} report the blended result. Replacing the baseline with the harness removes \TokD{} of token intensity and \CostD{} of cost while quality remains at parity, which is the operational signature of escaping token maxing: the token exchange rate improved rather than degraded. In efficiency terms, a dollar purchases \QPDgain{} more completed quality, and a million tokens completes \CPMgain{} more tasks. The hand-scored user-experience rubric moved in the same direction, for reasons that follow from the architecture: earlier streaming, live tool status in place of a spinner, and clean cancellation.

\subsection{Per-model results}
\label{sec:invariance}

\begin{figure}[t]
\centering
\includegraphics[width=\linewidth]{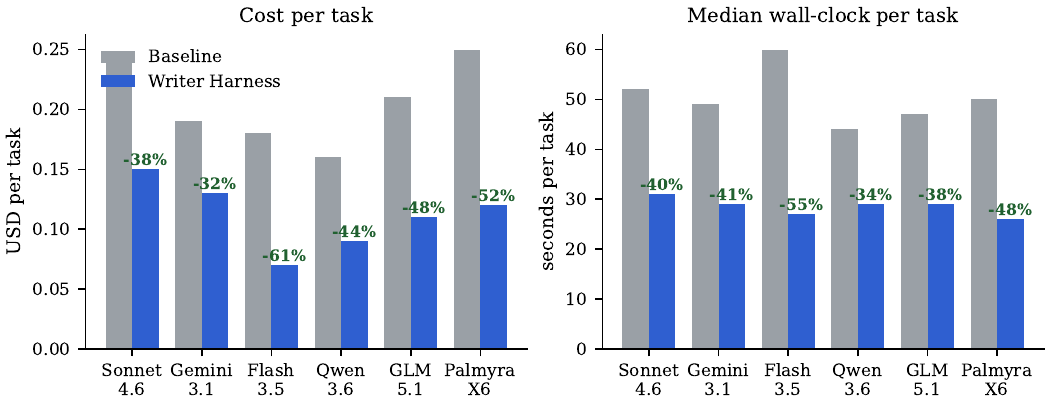}
\caption{Per-model cost and latency under the baseline and the harness. Cost and latency fall for every model; reductions range from $-$33\% to $-$61\% (cost) and from $-$33\% to $-$55\% (latency).}
\label{fig:permodel}
\end{figure}

\begin{table}[t]
\centering\small
\caption{Per-model efficiency, baseline $\rightarrow$ harness. Same tasks, same judges, same price table.}
\label{tab:permodel}
\begin{tabular}{@{}lcccccc@{}}
\toprule
 & \multicolumn{3}{c}{Cost / task} & \multicolumn{3}{c}{Wall-clock / task (median)} \\
\cmidrule(lr){2-4}\cmidrule(lr){5-7}
Model & Base & Harness & $\Delta$ & Base & Harness & $\Delta$ \\
\midrule
Claude Sonnet 4.6 & \$0.24 & \$0.15 & $-39\%$ & 52\,s & 31\,s & $-41\%$ \\
Gemini 3.1        & \$0.19 & \$0.13 & $-33\%$ & 49\,s & 29\,s & $-40\%$ \\
Gemini Flash 3.5  & \$0.18 & \$0.07 & $-61\%$ & 60\,s & 27\,s & $-55\%$ \\
Qwen 3.6          & \$0.16 & \$0.09 & $-44\%$ & 44\,s & 29\,s & $-33\%$ \\
GLM 5.1           & \$0.21 & \$0.11 & $-47\%$ & 47\,s & 29\,s & $-38\%$ \\
Palmyra X6        & \$0.25 & \$0.12 & $-52\%$ & 50\,s & 26\,s & $-48\%$ \\
\bottomrule
\end{tabular}
\end{table}

Table~\ref{tab:permodel} and Figure~\ref{fig:permodel} disaggregate the effect by model. Two observations follow. The first concerns uniformity: across six models, five vendors, and three weight classes, the cost of every model falls by at least one third. This is the signature of an effect at the level of the orchestration layer; savings arising from model-specific behavior would be visible in the spread. The second concerns magnitude relative to model selection. Under the baseline, moving from the most expensive model (Palmyra~X6 at \$0.25) to the least expensive (Qwen~3.6 at \$0.16) reduces cost by 36\%, whereas retaining any model and adopting the harness reduces it by \CostRange{}. On this workload, the orchestration layer is a larger cost lever than the choice of model.

The largest relative gains occur in the fast tier: Flash~3.5 falls by 61\% in cost and 55\% in latency. This is consistent with Eq.~\eqref{eq:decomp}. For small, inexpensive models, harness overhead (replayed history, broadcast schemas) constitutes a larger share of the total bill, so removing it removes proportionally more.

\subsection{Quality}
\label{sec:quality}

\begin{figure}[t]
\centering
\includegraphics[width=0.6\linewidth]{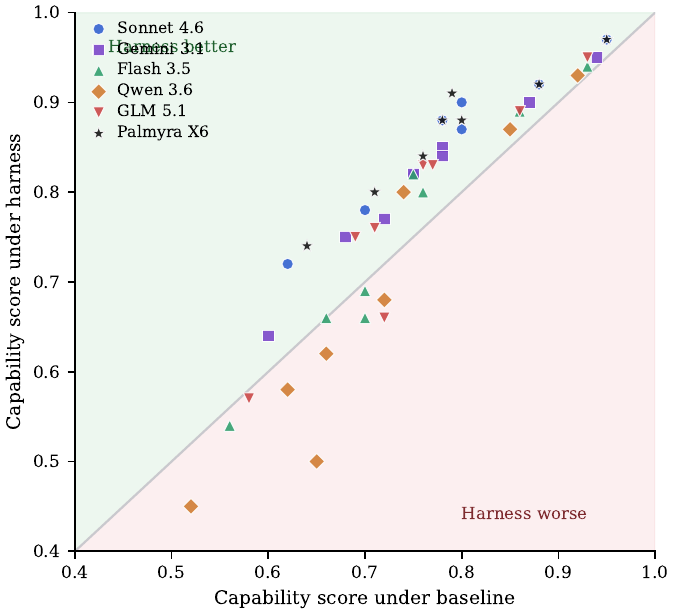}
\caption{Quality across the 48 capability$\times$model cells. Each point is one capability score for one model, baseline ($x$) against harness ($y$). Points above the diagonal improved under the harness: 30 improve, 11 are unchanged ($|\Delta|\le0.02$), and 7 regress. All regressions belong to the three smaller models and are concentrated in orchestration-intensive capabilities (MCP, Playbooks, Presentations).}
\label{fig:parity}
\end{figure}

Figure~\ref{fig:parity} plots all 48 capability$\times$model cells (eight legacy capabilities across six models; the full matrix appears in Appendix~\ref{app:matrix}). Most cells lie on or above the diagonal: 30 improve, 11 are unchanged, and 7 regress. The regressions are not randomly distributed. Each occurs on Flash~3.5, Qwen~3.6, or GLM~5.1, and six of the seven fall in the capabilities that exercise orchestration most heavily: MCP tool use (Qwen $-0.15$, GLM $-0.06$, Flash $-0.04$), Playbooks, and Presentations. The frontier models and Palmyra improve most in those same categories (MCP: Sonnet $+0.10$, Palmyra $+0.10$; GDR: Sonnet $+0.10$, Palmyra $+0.12$). The structure that a stronger model converts into quality appears to impose a load on a weaker one.

\subsection{Harness leverage}
\label{sec:leverage}

\begin{figure}[t]
\centering
\includegraphics[width=0.62\linewidth]{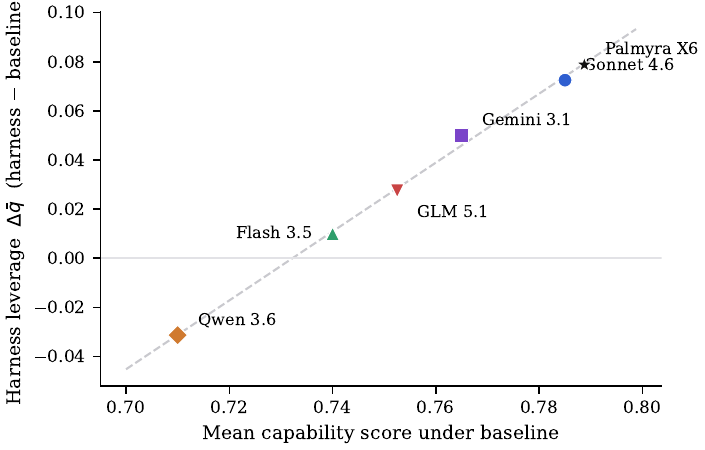}
\caption{Harness leverage as a function of baseline capability. Mean quality gain under the harness ($\Delta\bar q$ over eight capabilities) plotted against each model's mean baseline score. Stronger models obtain larger gains from the same orchestration layer ($r=0.99$; $n=6$, suggestive). Palmyra X6 ($+0.079$) and Sonnet 4.6 ($+0.073$) lead; Qwen 3.6 is net negative ($-0.031$).}
\label{fig:leverage}
\end{figure}

Averaging each model's scores over the eight capabilities yields a single measure of the quality it obtained from the change of orchestration layer: Palmyra~X6 $+0.079$, Sonnet~4.6 $+0.073$, Gemini~3.1 $+0.050$, GLM~5.1 $+0.028$, Flash~3.5 $+0.010$, and Qwen~3.6 $-0.031$. Plotted against the baseline score (Figure~\ref{fig:leverage}), the relationship is close to linear ($r=0.99$); with six points, it is suggestive rather than conclusive. We refer to the slope as \emph{harness leverage}, the rate at which a model converts orchestration structure into quality. The economic consequence is asymmetric. Efficiency gains are unconditional, whereas quality gains are contingent on capability: a weaker model under the harness still obtains its 44--61\% cost reduction but does not also improve. The two motivations for adopting a harness can therefore be assessed separately.

\subsection{Sub-agent delegation}
\label{sec:subagents}

\begin{table}[t]
\centering\small
\caption{Sub-agent delegation, a capability present only under the harness (no baseline arm exists). Scores are task-completion rates on delegation tasks, in which one or more scoped sub-agents are spawned and their results merged.}
\label{tab:subagents}
\begin{tabular}{@{}lcccccc@{}}
\toprule
 & Sonnet 4.6 & Gemini 3.1 & Flash 3.5 & Qwen 3.6 & GLM 5.1 & Palmyra X6 \\
\midrule
Sub-agents & 0.85 & 0.70 & 0.45 & 0.42 & 0.58 & 0.86 \\
\bottomrule
\end{tabular}
\end{table}

Delegation to spawned sub-agents, the one capability the harness introduces without a baseline counterpart, reaches a usable level of reliability only on the two strongest models (Table~\ref{tab:subagents}: Palmyra~X6 at 0.86, Sonnet~4.6 at 0.85), is degraded on Gemini~3.1 (0.70) and GLM~5.1 (0.58), and is not yet dependable in the fast tier (0.42--0.45). This is the harness-leverage result in its sharpest form: an orchestration feature carries a capability floor below which exposing it produces failures rather than function. Delegation also has a distinctive token profile. Sub-agent context is scoped, so delegated tokens are spent in a bounded side loop rather than inflating the parent context, under the context-firewall contract of Section~\ref{sec:mechanisms}: a bounded summary is returned, citations are carried in a sidecar, and the parent loop does not pay for the child's exploration.

\subsection{Per-prompt analysis}
\label{sec:cases}

Four exemplar prompts illustrate the distribution underlying the aggregates. The three-turn Medicare grounding dialogue improves from 0.60 to 0.90, the largest quality gain in the set, under a harness whose retrieval shaping supplies less but better-selected evidence, consistent with the finding that long, noisy contexts degrade attention \cite{lost}. The identity/refusal prompt holds at 0.90 in both arms (0.90/0.90) while its cost halves (\$0.04 to \$0.02). Contract Q\&A improves by $+0.07$ (0.75 to 0.82). The most expensive task in the set, multi-step research synthesis, falls from \$0.61 to \$0.33 ($-46\%$) but regresses in quality (0.80 to 0.60); this is the one case in which the aggregate parity conceals a real trade-off. The regression is driven by the smaller models, and it is the reason the deployment recommendation of Section~\ref{sec:discussion} withholds candidate models pending a fix rather than releasing with the regression. Secondary judges move with the headline measure: coherence from 0.85 to 0.88, communication from 0.79 to 0.80, and hallucination clean in both arms.

\section{Discussion}
\label{sec:discussion}

\subsection{The orchestration layer as price-setter}

The results separate along the lines of Section~\ref{sec:framing}. The terms of Eq.~\eqref{eq:decomp} that are determined by code (history, schemas, retrieval size, retries) moved for every model, because code executes identically regardless of which model reads its output. The quality terms are mediated by the model's capacity to exploit structure, and they moved with capability. The harness thus sets the price of work unconditionally and sets the ceiling on work jointly with the model. The layer that sets the price is also the layer through which an enterprise governs its agents; efficiency and control are concentrated in one component.

This bears on a common procurement practice. Teams that compare dollars per million tokens across vendors are comparing $p$; the bill is $p \times \tau$, and $\tau$ is determined by the harness. On this workload, the harness changed the bill by more than the full spread of the model menu (Section~\ref{sec:invariance}). An organization that rents its orchestration layer has delegated the variable over which it has the most control.

\subsection{Fleet-scale economics}

\begin{figure}[t]
\centering
\includegraphics[width=0.62\linewidth]{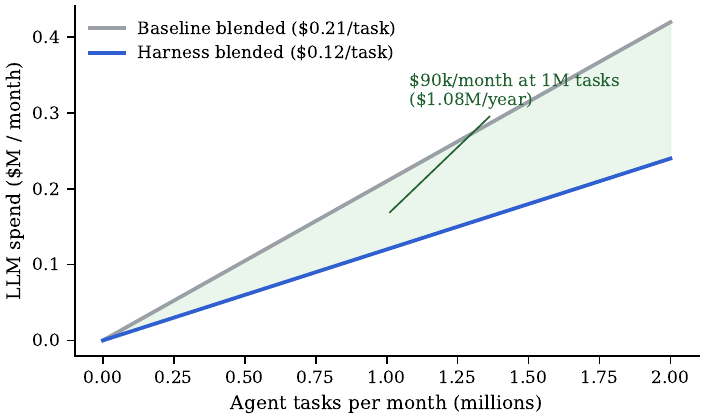}
\caption{Harness savings at fleet scale. Blended cost per task applied to monthly task volume. At one million agent tasks per month, the harness saves \$90k per month relative to the baseline (\$1.08M per year); the difference grows linearly with volume and applies to every model in the mix.}
\label{fig:fleet}
\end{figure}

Per-task differences understate the quantity of interest, which is fleet spend. At the blended rates measured here, an organization running one million agent tasks per month pays \$210k under the baseline loop and \$120k under the harness, a difference of \$1.08M per year attributable to the orchestration layer alone (Figure~\ref{fig:fleet}). Three properties distinguish this saving from other optimizations. It is model-portable: implemented above the model API, it held for all six models here and, by construction, applies to models that do not yet exist. It is linear in volume: it grows with agentic task volume, which is the fastest-growing quantity in enterprise deployments. And it is composable: declines in per-token prices \cite{epoch2025}, routing \cite{frugalgpt,routellm}, and prompt-level compression \cite{llmlingua,cod} multiply with it rather than substituting for it. Latency compounds in the same manner: a \SpeedX{} reduction in time per task is also a \SpeedX{} increase in throughput per unit of orchestration infrastructure, and shorter loops are less costly to retry.

\subsection{Implications for evaluation metrics}
\label{sec:kpi}

Definition~1 implies that the managerial remedy is a change in measurement. Teams that report quality alone have an incentive to token-max, because tokens appear on another budget line; teams that report $\mathrm{CPM}$ or $\eta_{\$}$ do not. The change of orchestration layer studied here moved $\mathrm{CPM}$ from \CPMA{} to \CPMB{} completions per million tokens, opposite to the prevailing trajectory, while quality held. We suggest that CPM should accompany quality in every agent release gate, for the same reason that performance per watt accompanies performance in processor design: it is the quantity that predicts the bill.

\subsection{Harness--model co-design and routing by feature demand}
\label{sec:codesign}

The pattern of regressions (all seven on smaller models, concentrated in MCP and Playbooks) and the sub-agent floor (Table~\ref{tab:subagents}) indicate that orchestration features are not costless to expose. A harness is a contract that the model must be capable of honoring. Two practical consequences follow. First, harness capabilities should degrade by model tier, reducing tool catalogs and disabling delegation below the floor, rather than presenting a single interface to all models. Second, routing research \cite{frugalgpt,routellm} should route on feature demand as well as on prompt difficulty: a request that will exercise sub-agents belongs on Palmyra~X6 or Sonnet~4.6 regardless of the apparent simplicity of its text, whereas a grounded question-answering request can be served by the fast tier at 61\% lower cost with no quality penalty (Figure~\ref{fig:parity}, GDR row: every model improves).

\subsection{Deployment implications}

Read as a deployment decision, the evidence supports general availability on the two models above the sub-agent floor (Palmyra~X6 and Sonnet~4.6), withholding the open-weight candidates pending the multi-step-research fix, and basing external claims on efficiency, which is uniform and decisive, rather than on quality, which is directional at $n=\NP{}$. We record this because it is the reading that Table~\ref{tab:scorecard} supports: the headline that can be defended is $-\TokD{}$ tokens at parity, not $+0.03$ quality.

\section{Threats to Validity}
\label{sec:threats}
\label{sec:limits}

\emph{Sample size.} A set of $n=\NP{}$ prompts is sufficient for the uniform and large efficiency differences but insufficient for inference about quality; all quality claims are directional, and the headline result is parity rather than improvement. \emph{Single-run baseline.} The baseline was run once and frozen (\FreezeDate{}); run-to-run variance in the baseline is not measured. \emph{Judge dependence.} Task completion is judged by language models; judge bias is partly mitigated by locked criteria and secondary judges but is not eliminated \cite{judge}. \emph{Price-table dependence.} Dollar figures inherit a single pinned public price table; token and latency results are independent of prices and lead to the same conclusions. \emph{Workload shape.} The task set reflects an enterprise assistant workload (grounding, workflows, tools, content); results may differ on long-horizon coding benchmarks \cite{swebench,sweagent}. \emph{Six points.} The harness-leverage correlation is computed over six models and should be read as a strong pattern that awaits confirmation on a wider panel. \emph{One pair.} We compare one baseline loop and one harness, both from a single vendor; the framing of Section~\ref{sec:framing} is general, but the magnitudes are specific to this pair. \emph{Cross-harness claims.} Section~\ref{sec:landscape} characterizes other systems from public documentation and a design-time source study; it contains no cross-harness measurements, and Table~\ref{tab:landscape} should be read as a comparison of designs rather than as a benchmark.

\section{Conclusion}

Held to the same \NP{} tasks and judged by the same panel, six foundation models completed the same work with \TokD{} fewer tokens, at \CostD{} lower cost, and in \LatD{} less time, because the software surrounding them changed and the models did not. The efficiency gains were unconditional; the quality gains accrued to the models capable of exploiting the additional structure; and the one new capability arrived with a capability floor. Token maxing is a choice made at the orchestration layer, and the mechanisms of Section~\ref{sec:mechanisms} show that it can be reversed there, by caching what is stable, compacting what is old, offloading what is large, suspending what is waiting, and bounding what is failing. On the evidence presented here, the harness is not ancillary infrastructure; it is the component that determines what an agent system costs.

\paragraph{Disclosure.} The authors are employed by Writer, Inc., which develops the Agent Harness and the Palmyra model family evaluated in this work; the last author is co-founder and CTO. The evaluation design (frozen baseline, locked prompts, identical judges and price tables across arms, and candidate-model failures scored rather than excluded) is intended to make the comparison auditable; the release artifacts are enumerated in Appendix~\ref{app:repro}.

\clearpage
\appendix

\section{Full Capability $\times$ Model Matrix}
\label{app:matrix}

\begin{table}[h]
\centering\small
\caption{Capability scores, baseline $\rightarrow$ harness, model held constant per cell. Codes: MSA = Model \& System Awareness; GDR = Grounding \& Retrieval; CNG = Content Generation; PLY = Playbooks; MCP = tool use via Model Context Protocol; PRN = Presentations (early-stage); VOX = Voice; IMG = Image Analysis \& Generation. Regressions ($\Delta < -0.02$) in bold.}
\label{tab:matrix}
\setlength{\tabcolsep}{4.5pt}
\begin{tabular}{@{}lcccccc@{}}
\toprule
Capability & Sonnet 4.6 & Gemini 3.1 & Flash 3.5 & Qwen 3.6 & GLM 5.1 & Palmyra X6 \\
\midrule
MSA & .95$\to$.97 & .94$\to$.95 & .93$\to$.94 & .92$\to$.93 & .93$\to$.95 & .95$\to$.97 \\
GDR & .80$\to$.90 & .78$\to$.85 & .75$\to$.82 & .74$\to$.80 & .76$\to$.83 & .79$\to$.91 \\
CNG & .80$\to$.87 & .78$\to$.84 & .76$\to$.80 & \textbf{.72$\to$.68} & .77$\to$.83 & .80$\to$.88 \\
PLY & .75$\to$.82 & .72$\to$.77 & .70$\to$.69 & \textbf{.66$\to$.62} & .71$\to$.76 & .76$\to$.84 \\
MCP & .78$\to$.88 & .75$\to$.82 & \textbf{.70$\to$.66} & \textbf{.65$\to$.50} & \textbf{.72$\to$.66} & .78$\to$.88 \\
PRN & .62$\to$.72 & .60$\to$.64 & .56$\to$.54 & \textbf{.52$\to$.45} & .58$\to$.57 & .64$\to$.74 \\
VOX & .88$\to$.92 & .87$\to$.90 & .86$\to$.89 & .85$\to$.87 & .86$\to$.89 & .88$\to$.92 \\
IMG & .70$\to$.78 & .68$\to$.75 & .66$\to$.66 & \textbf{.62$\to$.58} & .69$\to$.75 & .71$\to$.80 \\
\midrule
Mean & .785$\to$.858 & .765$\to$.815 & .740$\to$.750 & .710$\to$.679 & .752$\to$.780 & .789$\to$.867 \\
$\Delta\bar q$ & $+.073$ & $+.050$ & $+.010$ & $-.031$ & $+.028$ & $+.079$ \\
\midrule
Sub-agents (new) & .85 & .70 & .45 & .42 & .58 & .86 \\
\bottomrule
\end{tabular}
\end{table}

\section{Per-Prompt Exemplars and Secondary Judges}
\label{app:prompts}

\begin{table}[h]
\centering\small
\caption{Exemplar prompts (headline task-completion) and cost extremes. Secondary judges, blended: coherence .85$\to$.88; communication .79$\to$.80; hallucination clean in both arms. The remaining 18 prompts are reported in the per-prompt grid of the engineering evaluation report.}
\begin{tabular}{@{}lccl@{}}
\toprule
Prompt & Baseline & Harness & Note \\
\midrule
Medicare grounding (3-turn) & 0.60 & 0.90 & largest gain; retrieval shaping \\
Identity / refusal & 0.90 & 0.90 & parity; cost \$0.04$\to$\$0.02 ($-50\%$) \\
Contract Q\&A & 0.75 & 0.82 & gain \\
Multi-step research synthesis & 0.80 & 0.60 & \textbf{regression}; cost \$0.61$\to$\$0.33 ($-46\%$) \\
\bottomrule
\end{tabular}
\end{table}

\section{Derived-Metric Definitions}
\label{app:metrics}

Quality per dollar $\eta_{\$} = Q/C$ uses headline task-completion and blended cost per task: $0.78/\$0.21 = 3.71 \to 0.81/\$0.12 = 6.75$ ($+\QPDgain$). Completions per million tokens $\mathrm{CPM} = Q\cdot 10^{6}/\tau$: $0.78 \cdot 10^{6}/14{,}200 = \CPMA{} \to 0.81 \cdot 10^{6}/8{,}800 = \CPMB{}$ ($+\CPMgain$). Per-model harness leverage $\Delta\bar q$ averages the eight legacy capabilities of Appendix~\ref{app:matrix}; the sub-agent capability is excluded from means because it has no baseline arm.

\section{Reproducibility and Release Checklist}
\label{app:repro}

Artifacts to accompany the public release of this paper: (1)~the locked prompt set with pass/fail criteria; (2)~full per-turn traces (prompt tokens, completion tokens, tool events, \texttt{duration\_seconds}) for both arms and all six models; (3)~the pinned price-table snapshot used at report build; (4)~judge prompts and panel configuration; (5)~baseline and harness configuration manifests sufficient to identify the orchestration deltas of Section~\ref{sec:mechanisms}.

\end{document}